\documentclass{article} 
\usepackage{iclr2026_conference,times}

\usepackage{amsmath,amsfonts,bm}

\def\eqref#1{equation~\ref{#1}}

\def\1{\bm{1}}

\DeclareMathAlphabet{\mathsfit}{\encodingdefault}{\sfdefault}{m}{sl}
\SetMathAlphabet{\mathsfit}{bold}{\encodingdefault}{\sfdefault}{bx}{n}

\def\gE{{\mathcal{E}}}

\def\gP{{\mathcal{P}}}

\def\gS{{\mathcal{S}}}

\usepackage{pifont}
\usepackage{hyperref}
\usepackage{url}
\usepackage{enumitem}
\usepackage{booktabs}
\usepackage{todonotes}
\usepackage{wrapfig}
\usepackage{subcaption}
\usepackage{tcolorbox}
\usepackage{xcolor, colortbl}
\usepackage{multirow}
\usepackage{booktabs}
\usepackage{array}
\usepackage{adjustbox}
\usepackage{etoc}
\usepackage{graphicx}
\tcbuselibrary{skins, breakable}

\title{LinearRAG: Linear Graph Retrieval Augmented Generation on Large-scale Corpora}

\iclrfinalcopy

\author{}

\author{
{Luyao Zhuang$^{1}$}\thanks{Equal contribution. $^{\dagger}$Corresponding author.} , 
Shengyuan Chen$^{1*}$, 
Yilin Xiao$^{1}$, 
Huachi Zhou$^{1\dagger}$, Yujing Zhang$^{1}$\\
\textbf{ Hao Chen$^{1}$},
\textbf{Qinggang Zhang$^{1\dagger}$},
\textbf{Xiao Huang}$^{1}$\\
$^{1}$The Department of Computing, Hong Kong Polytechnic University, Hong Kong SAR\\
\texttt{\{luyao.zhuang,yilin.xiao,yu-jing.zhang\}@connect.polyu.hk}; \\
\texttt{huachi666.zhou@connect.polyu.hk}; \texttt{sundaychenhao@gmail.com};\\
\texttt{\{sheng-yuan.chen,qinggang.zhang,xiao.huang\}@polyu.edu.hk}
}

\begin{document}

\maketitle

\begin{abstract}

Retrieval-Augmented Generation (RAG) is widely used to mitigate hallucinations of Large Language Models (LLMs) by leveraging external knowledge. While effective for simple queries, traditional RAG systems struggle with large-scale, unstructured corpora where information is fragmented. Recent advances incorporate knowledge graphs to capture relational structures, enabling more comprehensive retrieval for complex, multi-hop reasoning tasks. However, existing graph-based RAG (GraphRAG) methods rely on unstable and costly relation extraction for graph construction, often producing noisy graphs with incorrect or inconsistent relations that degrade retrieval quality. In this paper, we revisit the pipeline of existing GraphRAG systems and propose \textbf{\underline{Linear}} Graph-based \textbf{\underline{R}}etrieval-\textbf{\underline{A}}ugmented \textbf{\underline{G}}eneration (\textbf{LinearRAG}), an efficient framework that enables reliable graph construction and precise passage retrieval. Specifically, LinearRAG constructs a relation-free hierarchical graph, termed Tri-Graph, using only lightweight entity extraction and semantic linking, avoiding unstable relation modeling. This new paradigm of graph construction scales linearly with corpus size and incurs no extra token consumption, providing an economical and reliable indexing of the original passages. For retrieval, LinearRAG adopts a two-stage strategy: (i) relevant entity activation via local semantic bridging, followed by (ii) passage retrieval through global importance aggregation. Extensive experiments on four datasets demonstrate that LinearRAG significantly outperforms baseline models. Our code and datasets are available at \textcolor{blue}{\url{https://github.com/DEEP-PolyU/LinearRAG.git}}.

\end{abstract}

\section{Introduction} 

Retrieval-Augmented Generation (RAG) has emerged as a promising approach to enhance Large Language Models (LLMs) by leveraging external knowledge bases~\citep{gao2023retrieval,lewis2020retrieval,zhou-etal-2025-taming,zhang2025faithfulrag}. However, existing RAG systems struggle with the complexities of large-scale, unstructured corpora in real-world scenarios, where the relevant information is frequently distributed unevenly across heterogeneous documents. The context retrieved by RAG systems is often voluminous, intricate, and lacks clear organization, leading to issues of variability in accuracy and coherence ~\citep{sun2024thinkongraph,zhang2024knowgpt}.
Although recent advances attempt to manage this by segmenting documents into smaller chunks for efficient indexing~\citep{borgeaud2022improving,izacard2023atlas,jiang2023active}, this strategy often results in the loss of critical contextual details, impairing retrieval accuracy and reasoning capabilities for complex tasks~\citep{han2024retrieval,zhang2025survey}.

To address this, Graph Retrieval-Augmented Generation (GraphRAG)~\citep{zhang2025survey,procko2024graph,xiao2025graphragbenchchallengingdomainspecificreasoning,xiang2025use} has recently emerged as a powerful paradigm that leverages external structured graphs to model the hierarchical structure of background knowledge~\citep{han2024retrieval}. Specifically, early work, like RAPTOR~\citep{sarthi2024raptor} and Microsoft's GraphRAG~\citep{edge2024local}, organizes knowledge through recursive summarization and community detection with LLM-generated synopses, enabling coarse-to-fine retrieval for comprehensive responses. 
Building on this, recent approaches, including GFM-RAG~\citep{luo2025gfm}, G-Retriever~\citep{he2024g}, and LightRAG~\citep{guo2024lightrag}, integrate specialized encoders and objectives, such as query-dependent GNNs, Prize Collecting Steiner Trees, and dual-level indexing, to improve multi-hop generalization, scalability, and efficiency. More recently, HippoRAG~\citep{hipporag} and its enhancement HippoRAG2~\citep{gutiérrez2025hipporag2} draw inspiration from cognitive processes to utilize personalized PageRank for multi-hop retrieval. These strategies significantly improve retrieval precision and contextual depth, enabling LLMs to address complex, multi-hop queries more effectively.

\begin{wrapfigure}{l}{0.46\textwidth}
\centering
\includegraphics[width=0.46\textwidth]{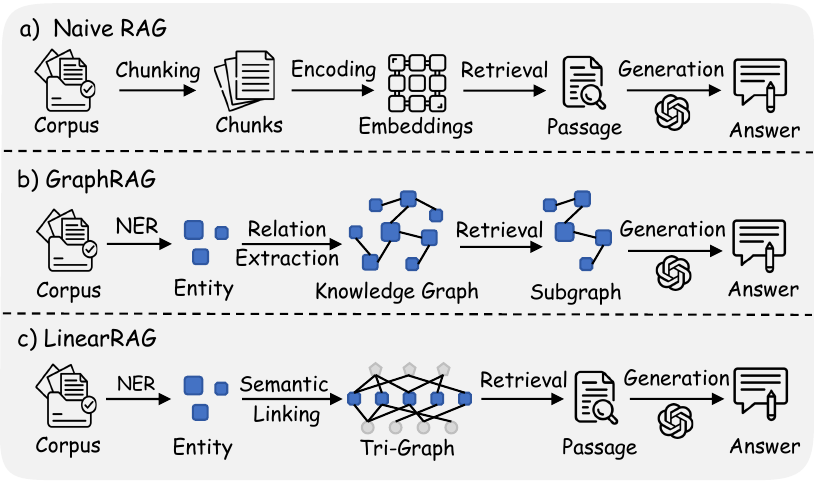}
\caption{Three paradigms of RAG systems.}
\label{fig:intro}
\vspace{-3mm}
\end{wrapfigure}
Despite its conceptual promise and theoretical superiority, recent studies reveal that GraphRAG models frequently underperform even naive RAG approaches on many real-world applications~\citep{han2025rag,zhou2025depth,xiang2025use,zhuang2025losemb}.
This performance degradation mainly stems from the poor quality of automatically constructed knowledge graphs. While graph-based retrieval increases recall of relevant knowledge, it concurrently introduces substantial noise and ambiguities into the retrieved contexts, due to errors in graph construction. Specifically, two critical deficiencies undermine graph quality, including (i) local inaccuracy: relation extraction processes exhibit significant error rates, resulting in inaccurate semantic relationships between entities. (ii) global inconsistency: the absence of mechanisms to enforce hierarchical consistency and global coherence during extraction leads to structurally fragmented graphs with poor connectivity. These deficiencies collectively manifest as structural conflicts and semantic ambiguity within the knowledge graph, which subsequently corrupt the retrieval and generation processes.
Although recent attempts have been made to refine graph quality via bottom-up clustering-based community summarization~\citep{edge2024local, gutiérrez2025hipporag2,wang2025archrag} or topic modeling~\citep{sarthi2024raptor} to offer a broader, macro-level view of data, these unsupervised methods are vulnerable to error propagation, where inaccuracies in entity relationships are amplified at higher levels of abstraction.

In this paper, we revisit the pipeline of existing GraphRAG systems and propose \textbf{\underline{Linear}} Graph-based \textbf{\underline{R}}etrieval-\textbf{\underline{A}}ugmented \textbf{\underline{G}}eneration (\textbf{LinearRAG}), a framework that enables efficient, reliable graph construction and precise corpus retrieval with multi-hop reasoning. The core idea of LinearRAG is to simplify the complex relational graph into a linear, easy-to-index view by focusing solely on modeling the semantics between target entities and the underlying text passages. Instead of relying on costly relation extraction, LinearRAG constructs a hierarchical graph from entities, sentences, and passages, using only lightweight entity extraction and semantic linking. On top of this graph, LinearRAG introduces a two-stage passage retrieval technique: \ding{182} \textbf{local semantic bridging for entity activation}, which identifies contextually relevant entities beyond literal matches by propagating semantic similarity in sentences to mine multihop contextual association; 
and \ding{183} \textbf{global importance aggregation for passage retrieval}, which applies personalized PageRank over the activated subgraph to aggregate passage importance from a holistic perspective. 
Together, these modules enable LinearRAG to achieve scalable, accurate, and noise-resilient retrieval for complex queries. Our overall contributions are summarized as follows:
\begin{itemize}
    \item We identify key limitations in existing GraphRAG systems, specifically highlighting how reliance on unstable relation extraction introduces noise and structural inconsistencies. This motivates the design of LinearRAG, a novel framework that enables reliable graph construction and precise passage retrieval while maintaining linear scalability.
    
    \item LinearRAG constructs a relation-free hierarchical graph, called Tri-Graph, using only lightweight entity extraction and semantic linking, which avoids the instability of traditional relation modeling and reduces indexing time by over $77\%$. 
    
    \item On top of the constructed graph, we design a two-stage retrieval mechanism that combines local semantic bridging for precise entity activation with global importance aggregation for passage recall. This integrated strategy enables more accurate, noise-resilient, and single-pass multi-hop retrieval.
    \item We conduct extensive experiments on four benchmark datasets, demonstrating that LinearRAG consistently outperforms state-of-the-art baselines in terms of retrieval quality, generation accuracy, and scalability, validating its practicality for real-world applications.
   
\end{itemize}

\section{Preliminary Study}
In this section, we conducted a series of preliminary studies to investigate the effect of graphs used in RAG systems. Our findings reveal critical flaws in graph construction that explain the underlying causes of GraphRAG's frequent underperformance compared to naive RAG in traditional tasks.

\begin{figure}[t]
  \centering
  \begin{subfigure}[t]{0.45\linewidth}  
    \centering
    \includegraphics[width=\linewidth]{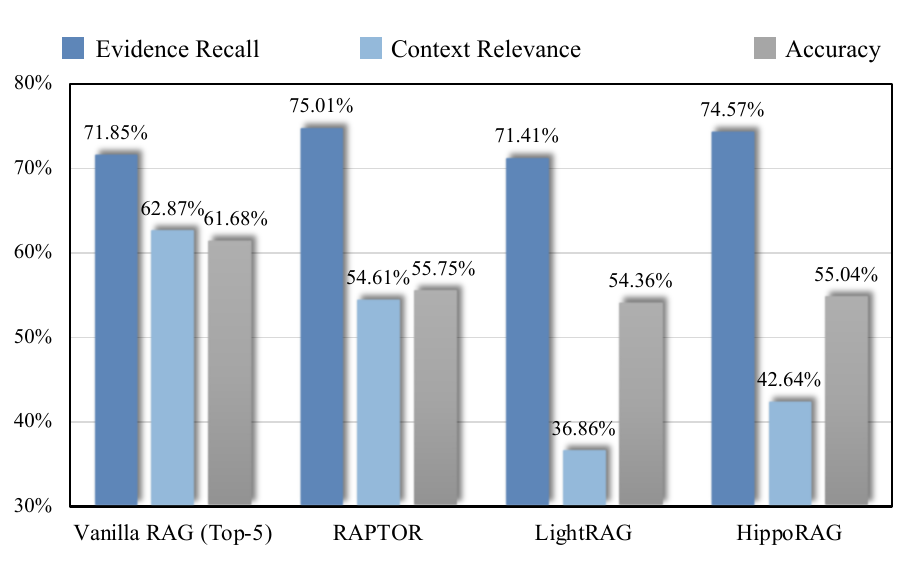}
    \caption{Retrieval and Generation Performance (\%) of vanilla RAG \textit{v.s.} typical GraphRAG baselines.}

    \label{fig:pre1}
  \end{subfigure}
  \hfill  
  \begin{subfigure}[t]{0.45\linewidth}  
    \centering
    \includegraphics[width=\linewidth]{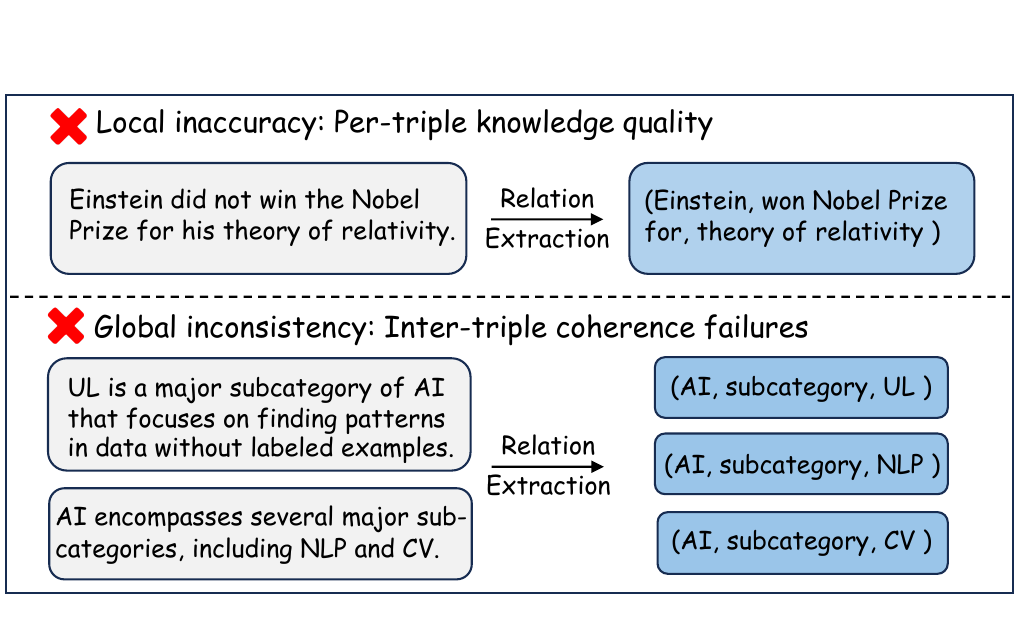}
    \caption{Two types of errors in knowledge graphs brought by imperfect relation extraction methods.}
    \label{fig:pre2}
  \end{subfigure}

\caption{\textbf{(a) Retrieval and generation rerformance (\%) of Vanilla RAG \textit{v.s.} GraphRAG Baselines.} Notably, the evaluation on Medical dataset measures GPT-based accuracy, context relevance, and evidence recall across different RAG baselines. \textbf{(b) Case study of relation errors in knowledge graph construction} from local inaccuracy and global inconsistency perspectives.}
\vspace{-6mm}

\label{fig:two_lines}

  \label{fig:two_lines}
\end{figure}

\subsection{Performance Degradation in GraphRAG Systems}
GraphRAG models frequently underperform traditional RAG approaches on many real-world tasks. Specifically, our experiments on GraphRAG-Bench~\citep{xiang2025use} demonstrate that while GraphRAG leverages graph structures to enhance recall by retrieving a wider array of potentially relevant passages, this gain is offset by a significant introduction of noisy and ambiguous contextual information. Specifically, GraphRAG methods such as LightRAG and HippoRAG achieve moderate improvements in evidence recall, but exhibit significantly lower context relevance, ranging from 36.86\% to 54.61\%, compared to Vanilla RAG, which attains 62.87\% performance. This suggests that although graph-based retrieval expands the scope of contextual information, it introduces substantial noise that compromises the relevance and reliability of generated answers. For example, in a question-answering task about ``climate change impacts'', GraphRAG might retrieve passages related to ``economic policies'' due to tenuous graph links. In contrast, vanilla RAG preserves tighter alignment with the query context, leading to more accurate and stable outputs.
\subsection{Graph Quality and Error Analysis}
To diagnose the root cause of performance degradation observed in Figure~\ref{fig:pre1}, we performed a fine-grained error analysis on the knowledge graphs used in GraphRAG. Our analysis revealed that this performance degradation stems directly from deficiencies in knowledge graph construction. Traditional GraphRAG pipelines rely on explicit relation extraction to construct relational graphs, which usually introduce errors at two levels:
(i) Local Inaccuracies: relation extraction models often produce factually incorrect triples. For example, as shown in Figure~\ref{fig:pre2}, the sentence ``Einstein did not win the Nobel Prize for his theory of relativity" may be misrepresented as (Einstein, won Nobel Prize for, theory of relativity), fundamentally altering factual meaning.
(ii) Global Inconsistencies:  existing relation extraction is performed locally on individual text passages, with no mechanism to validate or reconcile connections across the entire corpus, leading to redundant or contradictory relations. For example, ``AI" may be linked to ``Unsupervised Learning," ``NLP," and ``CV" as parallel subcategories without hierarchical coherence (e.g., that NLP and CV are subfields of AI, while ``Unsupervised Learning" is a technique used within them). This structural ambiguity directly misleads the retrieval process, introducing semantic noise based on these inconsistent connections.

\subsection{Discussion}
wThe conventional GraphRAG pipeline relies heavily on explicit relation extraction and triple-based knowledge representation. While this approach aims to summarize passages into structured relational forms, it faces two fundamental issues: \textbf{First, extracting concise and accurate relational triples is computationally expensive and linguistically challenging.} Relations expressed in natural language are often complex and context-dependent, and often nuanced or compositional to be accurately distilled into atomic triples; for example, the sentence ``Rachel reluctantly agreed to go running with Phoebe'' cannot be cleanly reduced to a single atomic triple without losing critical semantic nuance.
\textbf{Second, explicit relation extraction is unnecessary.} Aligned entities, rather than relations, serve as the primary anchors connecting information distributed across passages. The original text preserves relational semantics in full context, which can be interpreted dynamically by large language models during inference, without relying on error-prone extraction.

\begin{figure}[t]
    \centering
    \includegraphics[width=\linewidth]{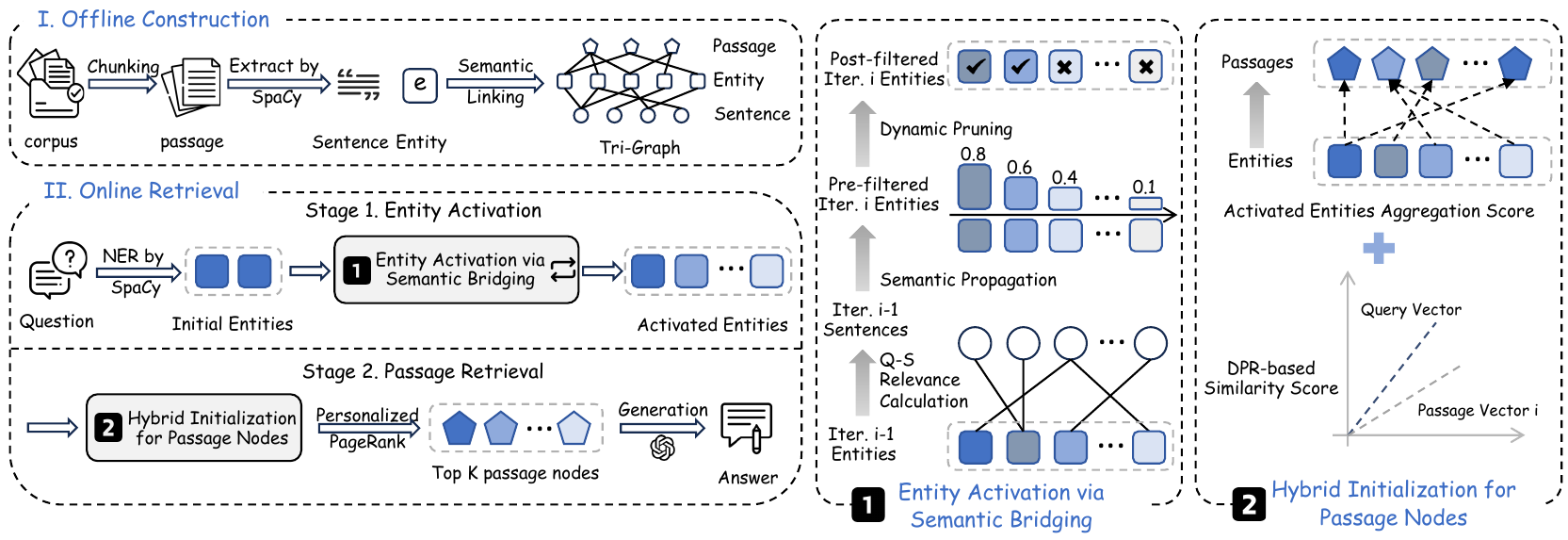}
    \caption{\textbf{The overall pipeline of the proposed LinearRAG framework.} \textbf{I. Offline Construction.} Initially, we construct a Tri-graph containing entity, sentence, and passage nodes, with edges connecting entities to sentences and entities to passages. \textbf{II. Online Retrieval.} We first activate relevant entities via local semantic bridging on the entity-sentence subgraph while fixing passage nodes, then using the activated entities to aggregate global importance scores, finally, perform passage retrieval via personalized PageRank on the entity-passage subgraph while fixing sentence nodes.}
    \label{fig:main_figure}
    \vspace{-7mm}
\end{figure}

\section{The Framework of LinearRAG}

Preliminary findings indicate that explicit relation extraction is not only computationally expensive but also largely unnecessary. These insights inform our revisit of the design of GraphRAG components, culminating in two central claims: (i) Aligned entities serve as the primary anchors connecting information distributed across passages. (ii) Contextual relations are best preserved within the original passages, eliminating the need for explicit relation extraction.

Motivated by this idea, we introduce a new GraphRAG paradigm as in Figure~\ref{fig:main_figure}. It constructs a graph that contains three types of nodes: entity nodes, sentence nodes, and passage nodes. Edges connect entities to sentences and entities to passages, represented as two adjacency matrices. Retrieval proceeds in two stages: \ding{182} \textbf{entity activation stage}, where we fix the passage nodes and use local semantic bridging on the entity–sentence subgraph to identify intermediate entities that connect different passages; and \ding{183} \textbf{passage retrieval stage}, where we fix the sentence nodes and apply personalized PageRank on the entity–passage subgraph, exploiting the activated entities in the first stage as seeds for global importance aggregation. This framework exhibits linear scalability in both graph construction and retrieval (detailed analysis in Appendix~\ref{complexity_analysis_appdx}), and we refer to it as \textbf{LinearRAG}.

\subsection{Token-free Graph Construction}\label{sec_g_construct}

We construct a hierarchical graph, called Tri-graph, with multiple granularities that is efficient to maintain and update. Given a corpus with a set of passages $\gP$, we first segment each passage into sentences using punctuation (e.g., periods or exclamation marks), obtaining a sentence set $\gS$. We then apply lightweight models (e.g., spaCy~\citep{honnibal2020spacy}) for named entity recognition (NER) to derive an entity set $\gE$. The passages, sentences, and entities constitute three types of nodes in the graph, denoted as $V_p$, $V_s$, and $V_e$, respectively.

Edges are constructed according to the following rules: if a passage $p_i$ contains an entity $e_j$, we add an edge $(V_{p_i}, V_{e_j})$; likewise, if a sentence $s_i$ mentions an entity $e_j$, we add an edge $(V_{s_i}, V_{e_j})$. These relations correspond to two adjacency matrices: the \textit{contain matrix} $C$ between passages and entities, and the \textit{mention matrix} $M$ between sentences and entities.  

Formally, the contain matrix $C$ is defined as an $|V_p| \times |V_e|$ matrix:
\begin{equation}
    C = [C_{ij}]_{|V_p| \times |V_e|}, 
    \quad \text{where} \quad 
    C_{ij} = \mathbb{1}_{\{p_i \text{ contains } e_j\}}. 
\end{equation}
Here, $\mathbb{1}$ denotes the indicator function, with $C_{ij}=1$ if $p_i$ contains $e_j$, and $C_{ij}=0$ otherwise.  

Similarly, the mention matrix $M$ is defined as an $|V_s| \times |V_e|$ matrix:
\begin{equation}
    M = [M_{ij}]_{|V_s| \times |V_e|}, 
    \quad \text{where} \quad 
    M_{ij} = \mathbb{1}_{\{s_i \text{ mentions } e_j\}}.
\end{equation}
This design enables efficient graph updates as the corpus grows. When new passages arrive, only those passages undergo sentence segmentation, NER, and edge construction, yielding overall linear complexity. Notably, NER is both precise and efficient compared with OpenIE, and can be performed using lightweight language models (e.g., BERT-based models in spaCy) without incurring LLM token costs. In addition, the adjacency matrices $C$ and $M$ are implemented in sparse form, exploiting their intrinsic sparsity to further reduce memory usage to a linear-scalability. Finally, by retaining the original passages as knowledge carriers, the resulting graph preserves all contextual information, thereby ensuring information-lossless construction.

\subsection{Passage Retrieval}\label{sec_psg_retrieval}
With the constructed graphs serving as lossless knowledge carriers, the objective becomes identifying the most informative passages effectively. The design of graph-based retrieval is critical, as it must balance precision and recall of relevant context, particularly in the case of multihop queries. We decompose the retrieval process into two stages: a precise entity activation stage via local semantic bridging followed by a passage retrieval stage to recall globally important passages.

\subsubsection{\textbf{First Stage}: Relevant Entity Activation via Semantic Bridging}\label{subsec_1st_stage}

Generally, direct entity matching can miss relevant intermediate entities that bridge multihop relations, which are essential for multihop queries. Therefore, it is crucial to identify these latent connectors. However, such intermediate entities cannot be directly mined, and the massive number of entities challenges the precision of identifying them. To tackle this, we propose \textit{relevant entity activation via semantic bridging}. For an incoming query \(q\), the process is as follows:


\textbf{Initial entity activation} that identifies entities contained in the query $q$ and represents their activation in the knowledge graph as a sparse vector. Specifically, we first use SpaCy to extract entities from $q$, denoted as $E_q$. Then for each extracted entity, we find the most similar entity in the knowledge graph. The initial activation score of the matched entity is set to its similarity score:

\begin{equation}
\mathbf{a}_q = [\mathbf{a}_{q,i}]_{|V_e| \times 1}, \quad \text{where} \quad \mathbf{a}_{q,i} = \mathbb{1}_{i=\arg\max_{e_j \in V_e} \texttt{sim}(e_q, e_j)} \cdot \texttt{sim}(e_q, e_i).
\end{equation}

\textbf{Query-sentence relevance distribution} that computes the contextual association between the query \(q\) and each sentence \( s_i \in S \), where \( S = \{s_1, s_2, \dots, s_{|S|}\} \) is the set of sentences in the corpus. Represent these similarities as a vector:
\begin{equation}
\mathbf{\sigma}_q = [\sigma_{q,i}]_{|S| \times 1}, \quad \text{where} \quad \sigma_{q,i} = \texttt{sim}(q, s_i).
\end{equation}
 \textbf{Semantic propagation} that propagates the similarities in \( \mathbf{\sigma}_q \) through semantic similarity extraction to activate relevant intermediate entities, enabling the bridging of multihop relations in the knowledge graph. The entity activation vector $ \mathbf{a}_q$ is updated with the weighted aggregation of their associated neighbors in the sentence-entity bipartite graph: 

\begin{equation}
   \mathbf{a}_q^{t} = \texttt{MAX}(M^T(\sigma_q \odot (M \mathbf{a}_q^{t-1})),\mathbf{a}_q^{t-1}),
   \label{eq:semantic_prop}
\end{equation}

where $\mathbf{a}_q^{t}$ is the activation vector of the entities in the $t_{th}$ iteration of semantic propagation. With a few iterations, we identify a set of contextually relevant entities that anchors a subgraph in the corpus that aligns with the reasoning structure of the query. This solution mimics the $\textit{relation-matching}$ process in the GraphRAG algorithms for multi-hop reasoning. Differently, our strategy performs implicit relation matching and does not rely on explicitly constructed relational knowledge graph. 

Noteworthy, the vectorized formulation of \eqref{eq:semantic_prop} enables $n$-hop entity activation within only $n$ iterations, where $n$ is generally small ($\leq4$). Each iteration involves three \texttt{MatMul} (matrix multiplication) and one \texttt{MAX} operation, which can be efficiently implemented and executed in parallel on popular machine learning platforms. Furthermore, due to the intrinsic sparsity of the matrix $M$ and the activation vector $\mathbf{a}_q^t$, we can store it in a sparse format to reduce memory consumption, and reduce computation cost by applying \texttt{SpMM} (sparse matrix multiplication) to replace \texttt{MatMul}.

\textbf{Dynamic pruning. }
While the above semantic bridging successfully establishes initial associations from query entities to potentially relevant intermediates, it also introduces a major challenge: the exponential growth of the search space as propagation deepens. Without proper control, irrelevant entities may repeatedly serve as new seeds, causing the process to expand combinatorially and drift into semantic regions unrelated to the original query intent. To address this issue, we perform graph pruning that constrains expansion to high-quality semantic paths. Specifically, at each propagation step, we introduce a threshold $\delta$. A newly activated entity is retained for the next iteration only if its relevance score exceeds $\delta$; otherwise, it is pruned. Moreover, the process terminates automatically once no new entity surpasses the threshold. This adaptive mechanism ensures that propagation proceeds only along the most relevant semantic paths, while dynamically tailoring the iteration range to the complexity of each query.

\subsubsection{\textbf{Second Stage}: Passage Retrieval via Global Importance Aggregation}
\label{subsec_2nd_stage}

In the first stage, relevant entities are extracted via local semantic-bridged subgraph expansion in the sentence-entity graph. These entities are used to initialize importance scores in a passage-entity graph, a bipartite graph where nodes represent passages (\( V_p \)) and entities (\( V_e \)), with edges connecting passages to their contained entities based on occurrence. The second stage performs global importance aggregation with hybrid initialization for passage nodes to retrieve important passages using Personalized PageRank (PPR) on the passage-entity graph to compute refined global importance scores for each node \( v_i \in V_p \cup V_e \):
\begin{equation}
    I(v_i) = (1-d) + d \cdot \sum_{v_j \in B(v_i)} \frac{I(v_j)}{\text{deg}(v_j)}
\end{equation}
Here, \( d \) is the damping factor (typically 0.85), \( B(v_i) \) is the set of nodes linking to \( v_i \), and \( \text{deg}(v_j) \) is the number of outgoing links from node \( v_j \). The initial importance score for entity node \( v_i \in V_e \) is set to \( I(v_i | v_i \in V_e) = \mathbf{a}_q^{(i)} \), where \( \mathbf{a}_q = (\mathbf{a}_q^{(1)}, \mathbf{a}_q^{(2)}, \dots) \) is a vector of entity relevance scores for query \(q\) computed in the first stage. For passage nodes \( v \in V_p \), the initial importance score is:
\begin{equation}
    I(v | v \in V_p) = \left( \lambda \cdot \texttt{sim}(q,v) + \ln\left(1 + \sum_{e_i \in E_a} \frac{\mathbf{a}_q^{(i)} \cdot \ln(1 + N_{e_i})}{L_{e_i}} \right) \right) \cdot W_p,
\end{equation}
where  \( \texttt{sim}(q,v) \) is the PPR base score capturing similarity between passage \( v \) and query \(q\), \( E_a \) is the set of activated entities from the first stage, \( N_{e_i} \) is the occurrence count of entity \( e_i \) in passage \( v \), \( L_{e_i} \) is the hierarchical level of entity \( e_i \), and \( W_p \) is the passage node weight coefficient, $ \lambda $ is a trade-off coefficient. Passages are ranked by their PPR scores \( I(v | v \in V_p) \), and the top-\( k \) passages with the highest scores are selected for retrieval.

\section{Experiments}

In this section, we conduct comprehensive experiments to verify the effectiveness and efficiency of LinearRAG. Specifically, we aim to answer the following questions. \textbf{Q1} (Generation Accuracy): How does LinearRAG perform compared to state-of-the-art GraphRAG methods in terms of generation performance? \textbf{Q2} (Efficiency Analysis): How cost-efficient and time-efficient is LinearRAG relative to existing GraphRAG approaches? \textbf{Q3} (Ablation Study): What contribution does each component of LinearRAG make to the overall performance? ( Note that for a comprehensive evaluation of LinearRAG, additional experiments on retrieval quality, parameter sensitivity, backbone analysis, large-scale efficiency analysis and case studies are presented in Appendix \ref{app:more_exp}.)

\begin{table*}[t]
    \caption{\textbf{Result (\%) of baselines and LinearRAG on four benchmark datasets in terms of both Contain-Match and GPT-Evaluation Accuracy.} The best result for each dataset is highlighted in \textbf{bold}, while the second result is indicated with an \underline{underline}.} 
    \label{tab:main_results}
    \small
    \centering
    \vspace{-3mm}
    \setlength{\tabcolsep}{1mm}
    \scalebox{0.96}{
    \begin{tabular}{lccccccc} 
    \toprule
    \multirow{2}{*}{\textbf{Method}}
        &\multicolumn{2}{c}{\textbf{HotpotQA}}
        & \multicolumn{2}{c}{\textbf{2Wiki}} 
        &\multicolumn{2}{c}{\textbf{MuSiQue}}
        & \textbf{Medical}\\
        \cmidrule(lr){2-3} \cmidrule(lr){4-5} \cmidrule(lr){6-7} \cmidrule(lr){8-8}
        &Contain-Acc. &GPT-Acc. &Contain-Acc. &GPT-Acc. &Contain-Acc. &GPT-Acc. &GPT-Acc. \\
    \midrule 
    \multicolumn{8}{c}{\textbf{\textit{Direct Zero-shot LLM Inference}}} \\
    \midrule
    
        llama-8B & 31.10 & 27.30 & 33.60 & 16.20 & 7.40 & 8.10 & 27.31 \\
        llama-13B & 24.20 & 16.80 & 21.90 & 10.50 & 3.30 & 4.40 & 28.86 \\
        GPT-3.5-turbo & 33.40 & 43.20 & 28.70 & 31.00 & 10.30 & 21.90 & 45.60 \\
        GPT-4o-mini & 38.90 & 40.20 & 36.30 & 31.40 & 13.60 & 15.80 & 42.10 \\
    \midrule
    \multicolumn{8}{c}{\textbf{\textit{Vanilla Retrieval-Augmented-Generation}}} \\
    \midrule
        Retrieval (Top-1) & 46.30 & 49.10 & 36.60 & 31.70 & 17.80 & 21.10 & 48.01 \\
        Retrieval (Top-3) & 53.00 & 56.00 & 44.90 & 39.70 & 25.10 & 27.50 & 59.07 \\
        Retrieval (Top-5) & 55.70 & 58.60 & 48.60 & 43.00 & 26.10 & 29.60 & 61.68 \\
    \midrule 
    \multicolumn{8}{c}{\textbf{\textit{Graph-based Retrieval-Augmented-Generation Methods}}} \\
    \midrule
        KGP & 61.50 & 60.90 & 31.60 & 30.00 & 25.60 & 30.10 & 54.22\\
 G-retriever& 42.20& 40.60& 46.60& 27.10& 14.40& 15.50&50.36\\
        RAPTOR & 55.90 & 58.30 & 50.10 & 42.10 & 23.30 & 27.40 & 55.75\\
        E$^2$GraphRAG & 61.00 & 63.90 & 54.30 & 38.10 & 23.80 & 26.20 & 58.00
\\
        LightRAG & 60.30 & 59.50 & 55.20& 39.00& 27.40& 28.60& 54.36\\
        HippoRAG & 57.00 & 59.30 & 66.10 & \underline{59.90} & 29.30 & 24.10 & 55.04
\\
        GFM-RAG & 62.70& \underline{65.60}& \underline{66.80} & 59.60 & 29.90 & 34.60 & 56.07\\
        HippoRAG2 & \underline{62.90} & 64.30 & 62.70 & 55.00 & \underline{31.00} & \underline{35.00} & \underline{60.77}\\
        \midrule
        
       \rowcolor[HTML]{cbddf5} \textbf{LinearRAG (Ours)} & \textbf{64.30} & \textbf{66.50} & \textbf{70.20} & \textbf{63.70} & \textbf{33.90} & \textbf{37.00} & \textbf{63.72}\\
    \bottomrule
    \end{tabular}}
    \vspace{-6mm}
\end{table*}

\subsection{Experimental Setting}

\textbf{Datasets.} We first evaluate the effectiveness of LinearRAG on three widely-used multi-hop QA datasets and one domain-specific dataset, including HotpotQA~\citep{yang2018hotpotqa}, 2WikiMultiHopQA (2Wiki)~\citep{2wikimqa}, MuSiQue~\citep{trivedi2022MuSiQue}, and the Medical dataset from GraphRAG-Bench~\citep{xiang2025use}. We follow the same evaluation method as HippoRAG, using the same corpus for retrieval and choosing 1,000 questions from each validation set. This setup allows for a fair comparison between different methods. We also test our approach on the domain-specific Medical dataset from GraphRAG-Bench~\citep{xiang2025use}, showing that LinearRAG improves both generation results and retrieval quality.

\textbf{Baselines.} We categorize all the baselines into three groups: (i) Zero-shot LLM Inference: We evaluate several foundational models including LLaMA3 (8B) and LLaMA3 (13B) ~\citep{dubey2024llama3}, as well as GPT-3.5-turbo and GPT-4o-mini~\citep{openai2023gpt4}. (ii) We deploy Vanilla RAG methodologies across multiple retrieval configurations (retrieving 1, 3, or 5 top passages), combining semantic-based document retrieval with chain-of-thought reasoning prompts to guide the LLM's generation process. (iii) State-of-the-art GraphRAG Systems: We compare against leading GraphRAG implementations including KGP ~\citep{wang2024knowledge}, G-retriever~\citep{he2024g}, RAPTOR ~\citep{sarthi2024raptor}, E$^2$GraphRAG~\citep{zhao20252GraphRAG}, LightRAG~\citep{guo2024lightrag}, HippoRAG~\citep{hipporag}, GFM-RAG~\citep{luo2025gfm}, and HippoRAG2~\citep{gutiérrez2025hipporag2}. Among these, RAPTOR organizes the corpus into a hierarchical tree graph; G-Retriever, LightRAG, HippoRAG, GFM-RAG, and HippoRAG2 extract triples from passages to build structured graphs; while E$^2$GraphRAG integrates both strategies.

\textbf{Evaluation Metrics.} We evaluate our method using four metrics across two categories. For end-to-end QA performance, following existing work (Wang et al., 2025), we use: 1) Contain-Match Accuracy (Contain-Acc.), which checks if the correct answer appears in the generated response, and 2) GPT-Evaluation Accuracy (GPT-ACC.), an LLM-based metric that assesses whether the predicted answer matches the ground truth. For the Medical dataset, since golden answers consist of lengthy descriptive statements, we only evaluate using GPT-ACC. For retrieval quality assessment, we adopt metrics from GraphRAG-Bench~\citep{xiang2025use}: 1) \textit{Context Relevance}, which measuring semantic alignment between questions and retrieved passages, and 2) \textit{Evidence Recall}, which evaluating whether the retrieved contents contain all necessary information for the correct answer.

\textbf{Implementations.} For consistency in implementation, all algorithms use the same embedding model, \textit{i.e.}, all-mpnet-base-v2~\citep{bge_embedding}). We set $k$ = 5 for top $k$ retrieval in all methods. All RAG approaches employ the same LLM (GPT-4o-mini) for both generation and evaluation tasks. All experiments are conducted on the hardware configuration detailed in Appendix~\ref{app:machine}.

\subsection{Generation Accuracy (Q1)}

To address Q1, we conduct a comprehensive evaluation of generation performance by comparing various baseline methods with LinearRAG across four benchmark datasets. The detailed experimental results are presented in Table \ref{tab:main_results}. Based on our analysis, we derive the following key observations.

\textbf{Obs. 1. RAG significantly enhances zero-shot LLM performance.} Direct prompting of LLMs without external knowledge retrieval yields the poorest performance across all datasets. For instance, advanced models like GPT-4o-mini achieve only 15.80\% GPT-based accuracy on the MuSiQue dataset when operating without retrieval augmentation. However, integrating relevant corpus content into prompts substantially improves performance, as Vanilla RAG with top-5 retrieved contexts increases accuracy to 29.60\% on the same dataset. Such significant performance enhancement demonstrates the essential nature of RAG mechanisms for information-intensive applications.

\textbf{Obs. 2.} \textbf{GraphRAG methods provide crucial context missed by RAG methods for complex reasoning.} While larger \textit{k} values improve accuracy, the gains diminish at higher values, revealing vanilla RAG's core limitation in multi-hop reasoning: it tends to overly focus on searching for mentioned entities or keywords within documents, thereby missing logic-related documents essential for the complete reasoning chain. By contrast, methods that model structural dependency in the retrieval consistently perform better. Among them, HippoRAG 2 achieves the best results across most datasets, attaining 62.90\% Contain-based accuracy on HotpotQA and 31.00\% on MuSiQue datasets.

\textbf{Obs. 3.} \textbf{LinearRAG demonstrates superior performance compared to GraphRAG methods, which suffer from sensitivity to constructed graph quality.} Typical GraphRAG methods address semantic misalignment by explicitly structuring knowledge into graphs, but their effectiveness relies heavily on relation extraction. Instead, LinearRAG gets rid of ineffective graph construction and LinearRAG outperforms all baselines by a significant margin across all datasets, empirically. Notably, LinearRAG achieves 63.70\% GPT-based accuracy on  2Wiki, around 3.80\% absolute improvement over the second best baseline, and boosts Contain-based accuracy to 70.20\%.

\begin{table*}[t]
   \caption{\textbf{Efficiency and performance comparison of different GraphRAG methods.} Notably, Accuracy represents the average of Contain-Acc. and GPT-Acc. metrics. Prompt tokens represent input to LLM, and completion tokens represent LLM output. Best results are highlighted in \textbf{bold}, and second-best results are \underline{underlined}.}
    \label{tab:efficiency}
    \centering
    \vspace{-3mm}
    \setlength{\tabcolsep}{2mm}
    \begin{tabular}{lccccc} 
    \toprule
    \multirow{2}{*}{\textbf{Method}}
        &\multicolumn{2}{c}{\textbf{Time (s)}}
        & \multicolumn{2}{c}{\textbf{Token Consumption ($\times10^6$)}} 
        & \multirow{2}{*}{\textbf{Accuracy}}\\
        \cmidrule(lr){2-3} \cmidrule(lr){4-5} 
        &Indexing &Retrieval (Avg.) &Prompt  &Completion  \\
         \midrule 
        G-retriever & 2745.94 & 11.487 & 6.05 & 2.26 & 36.85 \\
        RAPTOR & 1323.57 & \underline{0.062} & 0.81 & \underline{0.03} & 46.10 \\
        E$^2$GraphRAG & \underline{534.60} & \textbf{0.053} & \underline{0.78} & 0.08 & 46.20 \\
        LightRAG & 4933.22 & 10.963 & 35.52 & 51.16 & 47.10 \\
        HippoRAG & 936.00 & 1.461 & 3.05 & 0.98 & 63.00 \\
        GFM-RAG & 1202.77 & 1.211 & 3.05 & 0.98 & \underline{63.20} \\
        HippoRAG2 & 1147.01 & 1.694 & 4.98 & 1.22 & 58.85 \\
    \midrule
        \rowcolor[HTML]{cbddf5} \textbf{LinearRAG (Ours)} & \textbf{249.78}& 0.093& \textbf{0} & \textbf{0} & \textbf{66.95} \\
    \bottomrule
    \end{tabular}
    \vspace{-8mm}
\end{table*}

\subsection{Efficiency Analysis (Q2)}\label{sec_efficiency_exps}

To better understand the associated efficiency and cost implications, we conduct a dedicated analysis on prompt statistics across different GraphRAG models during the indexing and retrieval stages by comparing their token usage and running time on 2WikiMultiHopQA dataset. The results are presented in Table \ref{tab:efficiency}, and we summarize our key observations as follows:

\textbf{Obs. 4. Complicated graph construction inevitably brings significant time costs during both the indexing and retrieval stages.} Models such as G-Retriever and LightRAG employ intricate schema definitions, resulting in notably poor efficiency. For instance, LightRAG requires 4,933.22 seconds for indexing and 10.963 seconds for retrieval. These models define entities and keywords within the graph and perform both high-level and low-level retrieval. Consequently, LLMs are overburdened with processing the large volume of content retrieved from these two information sources, which results in higher fees and latency.

\textbf{Obs. 5. Reducing LLM token costs does not necessarily have a negative impact on performance.} 
Complicated prompt construction and completion in G-Retriever and LightRAG fail to deliver consistent performance improvements. For instance, HippoRAG2 uses only 3.05M tokens during prompt construction and 0.98M tokens in the completion stage, while HippoRAG2 consumes 4.98M tokens for prompt construction and 1.22M tokens in completion. Despite these lower token costs, both models outperform G-Retriever and LightRAG. 

\textbf{Obs. 6. LinearRAG delivers the strongest overall efficiency.} Considering the whole pipeline, LinearRAG is the fastest and introduces zero token usage. While E$^2$GraphRAG and RAPTOR are more time-efficient than LinearRAG at the retrieval stage, it comes at the cost of significantly reduced model performance, as it retrieves only directly related documents with the query without considering the structural dependencies in the reasoning chain. In comparison, the lightweight pipeline of LinearRAG avoids LLM calls for both indexing and retrieval, which minimizes latency and eliminates token costs. For deployments that demand strong performance along with speed, scalability, and cost control, LinearRAG is the most practical choice.

\subsection{Ablation Study (Q3)}

\begin{wrapfigure}{l}{0.5\textwidth}
\centering
\vspace{-5mm}
\includegraphics[width=0.49\textwidth]{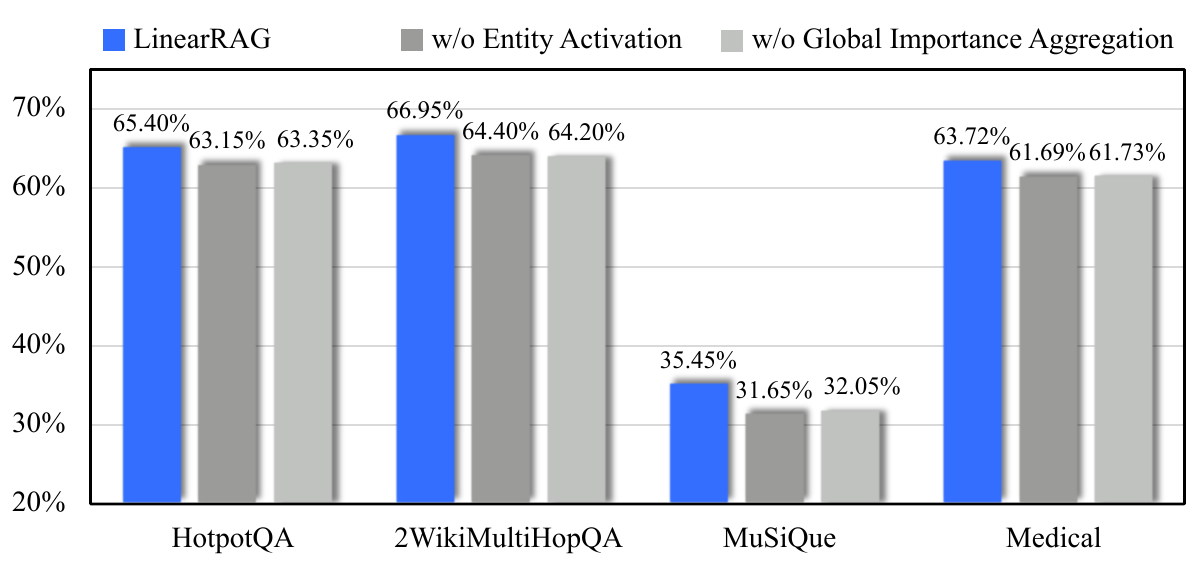}
\caption{\footnotesize \textbf{Ablation study on key modules of LinearRAG under four different datasets.} The y-axis represents the average of GPT-Acc. and Contain-Acc.}
\label{fig:ablation}
\vspace{-5mm}
\end{wrapfigure}

To address Q3, we conduct systematic ablation studies on the core components of LinearRAG across four datasets. We examine two key modules: \ding{182}  \textbf{Relevant Entity Activation via Semantic Bridging}, \textit{i.e., w/o Entity Activation}, which directly uses the initial entities extracted from the query as activated entities, bypassing the semantic bridging process that propagates activation scores from query entities to related entities in the knowledge graph. \ding{183} \textbf{Passage Retrieval via Global Importance Aggregation}, \textit{i.e., w/o Global Importance Aggregation}, which skips the personalized PageRank algorithm and directly uses the initial activation scores computed by stage \ding{182} to retrieve documents from the corpus, without considering global importance relationships. Each variant is evaluated with the average of GPT-Acc. \& Contain-Acc. as our primary evaluation metric. Experimental results are shown in Figure \ref{fig:ablation}, and we have the following findings.

\textbf{Obs. 7. } \textbf{Each module of LinearRAG is critical for optimal performance.} The performance gains are primarily attributed to two complementary stages. In the first Stage, LinearRAG identifies contextually relevant entities through semantic similarity propagation, uncovering the hidden logical relationships in multi-hop questions. In the second Stage, the model uses personalized PageRank algorithms on the activated subgraph to evaluate document importance from a global perspective. These two modules serve different but complementary purposes, allowing LinearRAG to achieve both effective and efficient performance.

\section{Conclusion}
In this work, we introduce LinearRAG, a novel GraphRAG framework that simplifies graph construction by replacing costly and error-prone relation extraction with lightweight entity extraction. It creates a hierarchical graph encompassing entities, sentences, and passages. Upon this, LinearRAG features a two-stage retrieval mechanism that advances the precision-recall Pareto frontier by jointly leveraging both local and global structural information. Extensive experiments demonstrate that LinearRAG consistently surpasses state-of-the-art baselines in retrieval precision, generation accuracy, and scalability, offering a robust and efficient solution for handling complex queries.

\section*{Ethics statement}
The first three datasets used in the experiments including HotpotQA,  2WikiMultiHopQA and MuSiQue are widely used datasets. The medical benchmark dataset is built from public resources. 
Our research strictly adheres to the ICLR Code of Ethics, particularly regarding data privacy, transparency,
and responsible computing practices. And there is no participant involved.
\section*{Reproducibility statement}
We provide detailed information required to reproduce the main experimental results of the paper, including data splits and hyperparameter configurations. Additionally, for each experiment, we outline the required computational resources, such as memory usage and execution time. All baseline models are sourced from public repositories. Our code and dataset are made available. 

\bibliography{iclr2026_conference}
\bibliographystyle{iclr2026_conference}
\appendix
\newpage

\section{Datasets}
Our experimental evaluation is conducted on four datasets: three established multi-hop benchmark datasets for multi-hop question answering—HotpotQA~\citep{yang2018hotpotqa}, MuSiQue~\citep{trivedi2022MuSiQue}, and 2WikiMultiHopQA (2Wiki)~\citep{2wikimqa}—and one domain-specific dataset. Below, we provide a concise overview of each dataset's key characteristics.

\textbf{(i) HotpotQA~\citep{yang2018hotpotqa}:} A comprehensive benchmark comprising 97k question-answer instances designed to evaluate multi-hop reasoning capabilities. Each question requires models to synthesize information from multiple documents, with up to 2 gold-standard supporting passages provided alongside numerous irrelevant documents. This structure challenges systems to perform effective cross-document inference and evidence selection.

\textbf{(ii) 2WikiMultiHopQA (2Wiki)~\citep{2wikimqa}:} A multi-hop reasoning benchmark containing 192k questions that necessitate information integration across multiple Wikipedia articles. Each instance requires evidence synthesis from either 2 or 4 specific articles, testing models' ability to perform structured cross-document reasoning and maintain coherent information flow.

\textbf{(iii) MuSiQue~\citep{trivedi2022MuSiQue}:} A sophisticated multi-hop QA benchmark featuring 25k question-answer pairs that demand 2-4 sequential reasoning steps. Each question requires coherent multi-step logical inference across multiple documents, challenging systems to execute structured reasoning chains while preserving contextual consistency throughout the inference process.

\textbf{(iv) Medical:} A specialized subset derived from GraphRAG-Bench~\citep{xiang2025use}, constructed from structured clinical data sourced from the National Comprehensive Cancer Network (NCCN) guidelines. These guidelines provide standardized treatment protocols, drug interaction hierarchies, and diagnostic criteria. The dataset encompasses four tasks of increasing complexity: fact retrieval, complex reasoning, contextual summarization, and creative generation, totaling 4,076 questions across all difficulty levels.

\section{Baseline Details}
In our experiments, we compare our method against several widely used GraphRAG approaches.

\textbf{(i) KGP}~\citep{wang2024knowledge} builds a knowledge graph over multiple passages with LLMs; at retrieval stage, it introduces an LLM-driven graph traversal agent to navigate the graph and progressively collect supporting passage.

\textbf{(ii) G-Retriever}~\citep{he2024g} combines graph neural networks with LLMs by formulating subgraph retrieval as a Prize-Collecting Steiner Tree optimization problem, enabling effective conversational question answering on textual graphs while mitigating hallucination and enhancing scalability.

\textbf{(iii) RAPTOR}~\citep{sarthi2024raptor} develops a hierarchical tree by applying clustering algorithms and abstractive summarization techniques, facilitating representation at multiple semantic granularities.

\textbf{(iv) E$^2$GraphRAG}~\citep{zhao20252GraphRAG} uses spaCy to extract entities and LLMs to summarize passage groups into a hierarchical tree with encoded nodes at indexing stage; at retrieval stage, it hits relevant entities in \(k\)-hop neighborhood and collects associated passages for ranking, otherwise dense retrieval over the whole tree.

\textbf{(v) LightRAG}~\citep{guo2024lightrag} employs a two-tier framework that incorporates graph-based representations within textual indexing, merging fine-grained entity-relation mappings with coarse-grained thematic structures.

\textbf{(vi) HippoRAG}~\citep{hipporag} is a training-free graph-enhanced retriever that uses the Personalized PageRank algorithm with query concepts as seeds for single-step or multi-hop retrieval across disparate documents.

\textbf{(vii) GFM-RAG}~\citep{luo2025gfm} implements a GraphRAG paradigm by constructing graphs from documents and using a graph-enhanced retriever to retrieve relevant documents.

\textbf{(viii) HippoRAG2}~\citep{gutiérrez2025hipporag2} extends HippoRAG with enhanced paragraph integration and contextualization. Optimizes seed node selection and PageRank reset probabilities while maintaining factual memory capabilities and improving associative memory performance. improving associative memory performance.

\section{Machine Configuration}
\label{app:machine}

All experiments in this study were conducted on the hardware configuration detailed in Table~\ref{tab:machine_config}. 

\begin{table}[h]
\centering
\caption{Detailed machine configuration used in our experiments.}
\label{tab:machine_config}
\begin{tabular}{lc}
\toprule
Component & Specification \\
\midrule
GPU & NVIDIA GeForce RTX 4090 D (24GB VRAM) \\
CPU & Intel(R) Xeon(R) Gold 6426Y \\
\bottomrule
\end{tabular}
\end{table}

\section{Efficiency Analysis of Graph Construction: \textbf{All-Stage Linear Scalability}}\label{complexity_analysis_appdx}

We present the efficiency analysis of LinearRAG and demonstrate that it achieves linear scalability in both runtime and memory consumption during graph construction and retrieval, thereby ensuring all-stage linear scalability. Detailed analysis is provided below.

\textbf{Graph construction stage} involves lightweight operations such as sentence segmentation and named entity recognition over the corpus. The resulting \textbf{computational complexity} is $O(|\gP|\cdot T)$, where $T$ denotes the average length of each passage. Notably, this stage incurs no LLM token consumption. Regarding memory usage, the first data structures to be stored are the embeddings of passages, nodes, and sentences, all of which scale linearly with the corpus size, i.e., $O(|\gP|\cdot T)$. The second set of data structures includes two adjacency matrices, $M$ and $C$, stored in sparse format by leveraging the inherent sparsity (each sentence contains at most $\sim4$ entities, and each passage at most $\sim10$ entities). Thus, memory consumption is $O(|\gP| + |\gS|)$, which simplifies to $O(|\gP|)$ since the number of sentences is proportional to the number of passages. Therefore, the \textbf{overall memory complexity} is $O(|\gP|\cdot T + |\gP|) = O(|\gP|\cdot T)$, i.e., linear with respect to corpus size.

\textbf{Retrieval stage} involves similarity computations, semantic propagation using \texttt{SpMM} (sparse matrix multiplication), and Personalized PageRank (PPR) iterations. Each propagation step requires $O(\text{nnz})$ time, where $\text{nnz}$ is the number of non-zero entries in the sparse matrices ($O(|\gS|)$), and PPR on the bipartite graph is computable in linear time with respect to the number of edges ($O(|\gP|)$). Overall, the \textbf{computational complexity} is $O(|\gP|)$. Memory consumption primarily arises from loading sparse graph structures, activation vectors (of size $O(|\gE| + |\gS| + |\gP|)$), and temporary query-specific computations.

\textbf{Acceleration with parallel computation.} The above analysis confirms the linear scalability of LinearRAG in both graph construction and retrieval. Additionally, due to the vectorized formulation of the graph structures, all data manipulations can be efficiently accelerated through parallel computation, further boosting performance in both stages. This is empirically verified in our efficiency experiments in section \ref{sec_efficiency_exps} and appendix \ref{appdx_large_corpus}.

\section{Additional Experiments}
\label{app:more_exp}
\subsection{Retrieval Quality Evaluation (Q4)}

To provide a more comprehensive assessment beyond generation performance, we adopt the settings from GraphRAG-Bench and utilize four tasks of increasing difficulty: Fact Retrieval, Complex Reasoning, Contextual Summarization, and Creative Generation. The comparison is conducted on the Medical dataset, same as GraphRAG-Bench. The results comparing LinearRAG with representative GraphRAG methods are presented in Table \ref{tab:retrieval_results} and we have the following observations.

\textbf{Obs. 8. GraphRAG models demonstrate significant improvements in recall metrics compared to RAG baselines, particularly in creative generation tasks, but fall behind in relevance metrics.} We infer that the reasons are twofold. On the one hand, for simpler tasks such as fact retrieval, models are not required to analyze reasoning chains to recall all the necessary supporting facts. Direct semantic retrieval is enough, reducing the risk of retrieving irrelevant or redundant information during graph traversal. On the other hand, for more complex tasks such as creative generation, the model needs continuous documents with strong logical dependencies to produce satisfactory outputs. RAG struggles to capture such complex relationships due to underlying graph structure of gold answers. Specifically, GFM-RAG improves recall from 44.88\% to 83.51\% in creative generation, but relevance drops from 58.73\% to 22.87\%.

\textbf{Obs. 9. LinearRAG demonstrates tremendous performance improvements over GraphRAG models in relevance metrics, owning the advantages of both high recall and strong relevance.} It is challenging to improve recall and relevance simultaneously because increasing recall often introduces more irrelevant documents, lowering relevance, while increasing relevance usually misses relevant results and lowering recall. LinearRAG achieves the best performance across four tasks in recall while achieves better than RAG in relevance metrics in most cases. We infer that the reason lies in our construction of a high-quality graph with minimal redundant connections, ensuring that irrelevant documents are not retrieved. Specifically, in complex reasoning tasks, LinearRAG achieves 87.03\% recall and 81.58\% relevance, significantly outperforming GFM-RAG's 85.03\% recall but only 33.06\% relevance, demonstrating LinearRAG's ability to maintain high precision while preserving comprehensive information coverage.

\begin{table*}[t]
    \caption{\textbf{Retrieval quality evaluation results (\%) of different baselines across four different question categories.} The table compares recall and relevance metrics for fact retrieval, complex reasoning, contextual understanding, and creative generation tasks. The best results are highlighted in \textbf{bold}, and second-best results are \underline{underlined}.} 
    \label{tab:retrieval_results}
    \small
    \centering
    \setlength{\tabcolsep}{1mm}
    \scalebox{0.96}{
    \begin{tabular}{lcccccccc} 
    \toprule
    \multirow{2}{*}{\textbf{Method}}
        &\multicolumn{2}{c}{\textbf{Fact Retrieval}}
        & \multicolumn{2}{c}{\textbf{Complex Reasoning}} 
        &\multicolumn{2}{c}{\textbf{Contextual}}
        & \multicolumn{2}{c}{\textbf{Creative Generation}}\\
        \cmidrule(lr){2-3} \cmidrule(lr){4-5} \cmidrule(lr){6-7} \cmidrule(lr){8-9}
        &Recall &Relevance &Recall &Relevance &Recall &Relevance &Recall &Relevance \\
    \midrule 
        Vanilla RAG (Top-5) & 86.24 & 63.71 & 84.97 & \textbf{84.11}& 84.14 & \textbf{89.94}& 44.88 & \underline{58.73}\\
        RAPTOR & 85.40 & 69.38 & \textbf{89.70}& 53.20 & 88.86 & 58.73 & 72.70 & 52.71 \\
        E$^2$GraphRAG & 87.84 & 69.74 & \underline{87.08}& 62.67 & \textbf{89.17}& 71.63 & 60.26 & 35.84 \\
        LightRAG & 80.32 & 41.27 & 82.91 & 42.79 & 85.71 & 43.11 & 81.34 & 45.17 \\
        HippoRAG & 87.25 & 52.44 & 83.80 & 42.19 & 83.46 & 49.13 & 81.66 & 45.03 \\
        GFM-RAG & \textbf{90.08}& 57.90 & 85.03 & 33.06 & 78.62 & 40.14 & \underline{83.51}& 22.87 \\
    \midrule
        \textbf{LinearRAG (Ours)} & \underline{88.86}& \textbf{86.09} & 87.03& \underline{81.58}& \underline{89.13}& \underline{87.89}& \textbf{89.08} & \textbf{72.74} \\
    \bottomrule
    \end{tabular}}
\end{table*}

\subsection{Hyper-parameter Sensitivity (Q5)}

\begin{wrapfigure}{r}{0.6\textwidth}
\centering
\begin{subfigure}[t]{0.29\textwidth}
  \includegraphics[width=\textwidth]{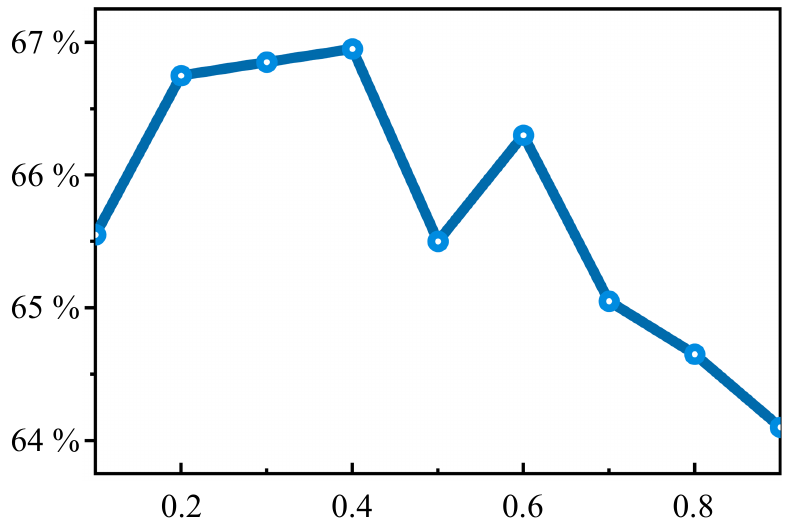}
  \caption{\footnotesize The impact of threshold $\delta$.}
  \label{fig:threshold}
\end{subfigure}
\hfill
\begin{subfigure}[t]{0.29\textwidth}
  \includegraphics[width=\textwidth]{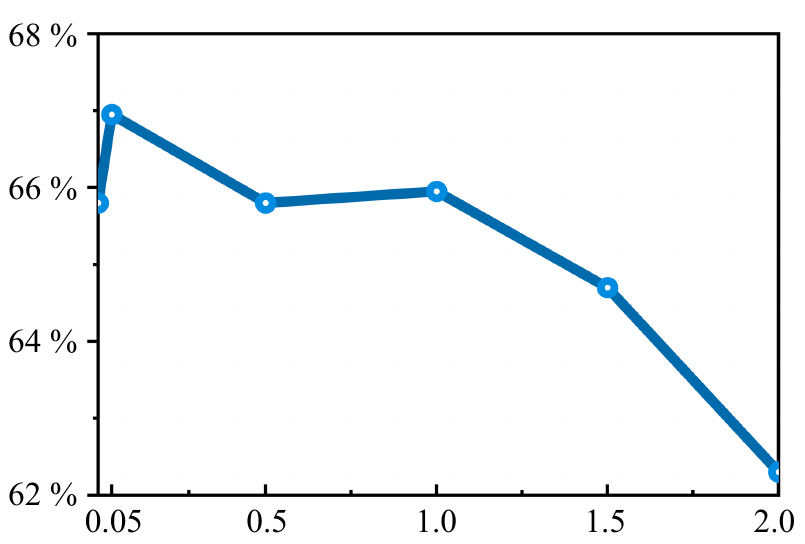}
  \caption{\footnotesize The impact of coefficient $ \lambda $.}
  \label{fig:ratio}
\end{subfigure}
\caption{\footnotesize \textbf{Parameter analysis of LinearRAG performance in the 2WikiMultiHopQA dataset.} (a) shows the dependency of LinearRAG performance on threshold $\delta$ in dynamic pruning. (b) examines the effect of trade-off coefficient $ \lambda $.}
\label{fig:parallel}
\vspace{-3mm}
\end{wrapfigure}

We conduct ablation studies to investigate the impact of key hyperparameters in LinearRAG. All experiments are evaluated using the average scores of  of GPT-Acc. and Contain-Acc. on 2WikiMultiHopQA.

\textbf{Obs. 10. Impact of threshold $\delta$.} The threshold $\delta$ determines whether an entity's score is retained during dynamic pruning and whether to continue expansion. As shown in Figure~\ref{fig:threshold}, when $\delta$ is too small, it introduces excessive noise and reduces efficiency in the retrieval stage. However, when $\delta$ is too large, it prevents the incorporation of more relevant entities, limiting the model's ability to capture comprehensive contextual information. Therefore, selecting an appropriate value for $\delta$ is crucial for balancing retrieval quality and computational efficiency. And, we set $\delta = 4$ based on our experimental setting.

\textbf{Obs. 11. Impact of trade-off coefficient $\lambda$.} The parameter $\lambda$ controls the balance between DPR-based passage similarity and entity-level information during passage initial scoring. As observed in Figure~\ref{fig:ratio}, optimal performance is achieved when $\lambda$ takes a relatively small value (e.g., 0.05), ndicating that entity information serves as the primary component while DPR similarity score acts as an auxiliary enhancement.

\subsection{Effectiveness of Different Sentence Embeddings (Q6)}

\begin{table*}[t]
    \caption{\textbf{Comparison of different sentence embedding models used in LinearRAG.} The best results are highlighted in \textbf{bold}.} 
    \label{tab:embedding_model}
    \small
    \centering
    \setlength{\tabcolsep}{1mm}
    \begin{tabular}{lccccccc} 
    \toprule
    \multirow{2}{*}{\textbf{Model}}
        &\multicolumn{2}{c}{\textbf{HotpotQA}}
        & \multicolumn{2}{c}{\textbf{2Wiki}} 
        &\multicolumn{2}{c}{\textbf{MuSiQue}}
        & \textbf{Medical}\\
        \cmidrule(lr){2-3} \cmidrule(lr){4-5} \cmidrule(lr){6-7} \cmidrule(lr){8-8}
        &Contain-Acc. &GPT-Acc. &Contain-Acc. &GPT-Acc. &Contain-Acc. &GPT-Acc. &GPT-Acc. \\
    \midrule 
    all-mpnet-base-v2 & 64.30& 66.50& \textbf{70.20}& \textbf{63.70}& \textbf{33.90}& \textbf{37.00}& 63.72\\
    all-MiniLM-L6-v2& 64.20& 64.90& 69.50& 62.60& 32.80& 36.90& 62.32\\
    bge-large-en-v1.5& 66.20& 67.60& 69.80& 63.90& 31.80& 35.20& 64.45\\
    e5-large-v2& \textbf{66.80}& \textbf{68.30}& 69.90& 63.10& 31.70& 36.50& \textbf{65.42}\\
    \bottomrule
    \end{tabular}
\end{table*}

In this section, we study the effectiveness of different sentence embeddings in the LinearRAG. We compare the all-mpnet-base-v2~\citep{song2020mpnet}, bge-large-en~\citep{bge_embedding}, all-MiniLM-L6-v2~\citep{wang2020minilm} and e5-large-v2~\citep{wang2022text}. We download the official pre-trained model from the Huggingface2. 

\textbf{Obs. 12.} Based on the findings presented in Table \ref{tab:embedding_model}, we observe that performance variations across different sentence embedding models remain relatively modest, with all-mpnet-base-v2 demonstrating superior results on the majority of datasets. This suggests that LinearRAG exhibits robustness to the selection of sentence embedding models. Consequently, we adopt all-mpnet-v2 as our default embedding models in all experiments due to its computational efficiency.

\begin{table*}[t]
\caption{\textbf{Indexing efficiency comparison of GraphRAG methods on large-scale ATLAS-Wiki corpus.} Results are evaluated on ATLAS-Wiki subsets of 5M and 10M tokens. Notaly, prompt tokens represent input to LLM, and completion tokens represent LLM output.}
    \label{tab:large}
    \centering
    \vspace{-3mm}
    \setlength{\tabcolsep}{2.5mm}
    \begin{tabular}{lcccc} 
    \toprule
    \multirow{2}{*}{\textbf{Dataset}} & \multirow{2}{*}{\textbf{Method}} & \multirow{2}{*}{\textbf{Indexing Time (s)}} & \multicolumn{2}{c}{\textbf{Token Consumption ($\times10^6$)}} \\
    \cmidrule(lr){4-5} 
    & & & \textbf{Prompt} & \textbf{Completion} \\
    \midrule 
    \multirow{3}{*}{5M} & RAPTOR & 18033.75 & 7.43 & 0.96 \\
    & HippoRAG & 6032.46 & 13.94 & 3.68 \\
    & \textbf{LinearRAG (Ours)} & \textbf{1409.95} & \textbf{0} & \textbf{0} \\
    \midrule
    \multirow{3}{*}{10M} & RAPTOR & 46430.96 & 16.62 & 2.82 \\
    & HippoRAG & 13815 & 28.04 & 7.53 \\
    & \textbf{LinearRAG (Ours)} & \textbf{3084.38} & \textbf{0} & \textbf{0} \\
    \bottomrule
    \end{tabular}
    \vspace{-6mm}
\end{table*}

\subsection{Large-Scale Efficiency Analysis (Q7)}\label{appdx_large_corpus}

To evaluate the scalability and effectiveness of LinearRAG on large corpora, we conduct comprehensive efficiency analysis using the ATLAS-Wiki dataset~\citep{bai2025autoschemakg}. We create two subsets containing 5M and 10M tokens respectively for indexing phase evaluation, as shown in Table \ref{tab:large}.

\textbf{Obs. 13.} The results demonstrate that LinearRAG achieves significant efficiency advantages over typical GraphRAG methods. Specifically, LinearRAG eliminates token consumption entirely (0 prompt and completion tokens) while achieving 12.8× and 15.1× speedup compared to RAPTOR on 5M and 10M datasets respectively. This zero-cost indexing approach makes LinearRAG particularly suitable for large-scale enterprise deployments without API dependencies.

\subsection{Case Study (Q8)}

\begin{table*}[h]
\caption{\textbf{Case study of LinearRAG.} LinearRAG successfully captures implicit relationships without requiring explicit relation extraction, enabling correct multi-hop reasoning.}
\label{tab:case}
\small
\centering
\resizebox{0.98\textwidth}{!}{%
\begin{tabular}{p{3cm}|p{13cm}}
\toprule
\textbf{Question} & "What nationality is Beatrice I, Countess Of Burgundy's husband?" \\
\midrule
\textbf{Ground Truth} & Germany \\
\midrule
\textbf{Support Context} & ["Beatrice I, Countess of Burgundy" $\to$ "spouse" $\to$ "Frederick Barbarossa"] \\
& ["Frederick Barbarossa" $\to$ "country of citizenship" $\to$ "Germany"] \\
\midrule
\textbf{Hippo2RAG} 
& \textbf{\textit{Retrieved context:}}\\&
1) \ding{51} ``William Duncan (actor)'': ...frederick barbarossa was elected king of germany at frankfurt on 4 march 1152...\\&
2) \ding{53} ``Dulce of Aragon'': "Dulce of Aragon (or of Barcelona; ; 1160 \u2013 1 September...\\&
3) \ding{53} ``Gordon Flemyng'': Gordon William Flemyng( 7 March 1934 \u2013 12 July 1995) was a...\\&
4) \ding{53} ``Marian Hillar'': Marian Hillar is an American philosopher, theologian, linguis...\\&
5) \ding{53} ``Semi-Tough'': Semi-Tough is a 1977 American comedy film directed by Michael...\\&
\textbf{\textit{Prediction:}}\\&
\ding{53} French. \\
\midrule
\textbf{LinearRAG} & 
\textbf{\textit{Activated Entities:}}\\&
Tter 0-Entity: Beatrice I \\&
Tter 0-Sentence: Beatrice I (1143 \u2013 15 November 1184) was Countess of Burgundy from 1148 until her death, and was also Holy Roman Empress by marriage to Frederick Barbarossa. \\&
Tter 1-Entity: Frederick Barbarossa \\&
Tter 1-Sentence: Frederick Barbarossa was elected King of Germany at Frankfurt on 4 March 1152 and crowned in Aachen on 9 March 1152. \\&
Tter 2-Entity: Germany \\&

\textbf{\textit{Retrieved context:}}\\&
1) \ding{51} ``William Duncan (actor)'': ...he was elected king of germany at frankfurt on 4 march 1152...\\&
2) \ding{51} ``Beatrice I, Countess of Burgundy'': ...holy roman empress by marriage to frederick barbarossa...\\&
3) \ding{53} ``Richilde, Countess of Hainaut'': Richilde, Countess of Mons and Hainaut( \u2013 15 March 1086)...\\&
4) \ding{53} ``Sophie of Pomerania, Duchess of Pomerania'': Sophia of Pomerania- Stolp( 1435 \u2013 24...\\&
5) \ding{53} ``"D\u00e1ire Drechlethan'': "D\u00e1ire Drechlethan\" D\u00e1ire of the Broad Face\" is a...\\&
\textbf{\textit{Prediction:}}\\&
\ding{51} Germany. 
 \cr \bottomrule
\end{tabular}}
\end{table*}

To clearly contrast typical GraphRAG baselines with our LinearRAG framework, we include a detailed case analysis in Table \ref{tab:case}, comparing results from the strong baseline HippoRAG2 and our model on a multi-hop question from the 2WikiMultihopQA dataset.

\textbf{Obs. 14.} The case demonstrates that HippoRAG2 fails to retrieve relevant evidence due to its dependence on explicitly pre-extracted relations (such as husband) that are missing in this context. In comparison, our LinearRAG method effectively captures implicit relationships through entity activation and contextual chaining, without dependence on pre-extracted relational tuples.

\section{Related Work}

Large language models are prone to hallucination~\citep{fang2024alphaedit,jiang2025anyedit,fang2025safemlrm,zheng2025usb,hong2024next,yuan2025knapsack,zhong2024iterative}, while RAG is a promising solution by grounding the reasoning process on contextual evidence from knowledge bases~\citep{zhang2025erarag,zhou2025depth,xiao2025graphragbenchchallengingdomainspecificreasoning,zhang2025survey,xiang2025use}.
However, real-world knowledge is often distributed across documents, organizing them effectively for answering complex questions has always been a challenging yet promising research topic in RAG. In the following, we discuss two major lines (\ding{182}, \ding{183}) of graphRAG research that closely relate to our work, which construct explicit graphs for organizing external knowledge sources. Then discuss reasoning-enhanced RAG (\ding{184}), which leverages LLM's inherent reasoning ability for answering complex questions without explicitly structuring corpora.

\ding{182} \textbf{Clustering-based hierarchy construction.} One line of research employs clustering-based community detection, a bottom-up method that applies algorithms like Louvain or Leiden to identify densely connected clusters of entities within the initial graph~\citep{edge2024local,sarthi2024raptor, hipporag}. By grouping related entities, it creates a hierarchical structure that abstracts passages into topic-based communities, reducing redundancy and offering a broader, macro-level view of the data. However, as an unsupervised technique, it is vulnerable to error propagation, where inaccuracies in entity relationships are amplified at higher levels of abstraction. Moreover, applying these clustering algorithms to large-scale graphs presents significant scalability issues, making real-time applications infeasible in practice.

\ding{183} \textbf{Relation-extraction-based knowledge graph construction.} This line of research~\citep{hipporag,guo2024lightrag,xiao2025reliablereasoningpathdistilling} adopts the idea of knowledge graphs to organize knowledge across different passages. The basic idea is to extract triples from each text chunk (passage) as an atomic summarization of the knowledge within the passage. The triples are connected via entity alignment~\citep{neusymea, LLM4EA}, ultimately forming a unified knowledge graph, which directly carries structured knowledge and serves as the index of passages, where off-the-shelf graph reasoning algorithms~\citep{difflogic, pLogicNet, liu2023rsc} can be applied to perform reasoning~\citep{sun2024thinkongraph,luo2024reasoning}. However, due to context window limits, OpenIE on each passage is processed independently, so the generated triples may be inconsistent. Recent work~\citep{liang2024kag,sharma2024og} mitigates this via top-down graph construction guided by a globally defined schema, but it relies on manual expert annotation to create and maintain the domain-specific schema, which is not only expensive, time-consuming, and does not generalize well across domains.

\ding{184} \textbf{Reasoning-enhanced RAG.} The key to answering complex questions in RAG is to effectively identify multiple distributed documents that support multihop logical dependencies. While GraphRAG methods explicitly construct graphs with human priors to organize corpora for efficient structured retrieval, reasoning-enhanced RAG directly leverages the inherent reasoning ability of LLMs to decompose complex queries into simpler subquestions that can be addressed by dense retrieval. LogicRAG~\citep{chen2025logicrag} and LAG~\citep{xiao2025lag} formalize this idea by first decomposing a query into a set of subqueries, then constructing a directed acyclic graph based on the logical dependencies among them, and solving subproblems one by one by following the topological order. Similarly, Chain-of-Note (CoN)~\citep{chainofNote} enhances RAG by generating sequential notes that break down complex queries into intermediate reasoning steps, retrieving relevant documents for each step to build a coherent answer. Another approach, SelfRAG~\citep{asai2024selfRAG}, integrates self-reflection into the RAG process, where the LLM iteratively evaluates and refines subqueries to ensure retrieved documents align with the logical flow of the reasoning process. These methods collectively highlight the potential of LLM-driven reasoning to improve retrieval efficiency and answer quality for complex, multihop queries.

\section{The Use of Large Language Models}
We employ LLMs primarily for writing polish, including correcting spelling errors, fixing grammatical issues, and rewriting non-native expressions to improve clarity and fluency. And LLMs are used only for writing polishing of our manuscript and appendix. It does not generate research ideas, results, or claims. We ensure that all scientific contributions and implementations are original. 

\end{document}